\documentclass[review]{elsarticle}

\usepackage{lineno,hyperref}
\usepackage{mathtools}
\usepackage[algo2e]{algorithm2e}
\usepackage{algorithm}
\usepackage{pgfplots}
\usepackage{tabularx}
\usepackage{longtable}
\usepackage{verbatim}
\modulolinenumbers[5]
\newcommand{\specialcell}[2][c]{%
	\begin{tabular}[#1]{@{}c@{}}#2\end{tabular}}

\makeatletter
\def\ps@pprintTitle{%
	\let\@oddhead\@empty
	\let\@evenhead\@empty
	\let\@oddfoot\@empty
	\let\@evenfoot\@oddfoot
}
\makeatother

\biboptions{authoryear}

\begin{document}

\begin{frontmatter}

\title{CNNPred: CNN-based stock market prediction using several data sources}

\author[mymainaddress]{Ehsan Hoseinzade}
\ead{hoseinzadeehsan@ut.ac.ir}
\author[mymainaddress]{Saman Haratizadeh}
\ead{haratizadeh@ut.ac.ir}

\address[mymainaddress]{Faculty of New Sciences and Technologies, University of Tehran, North Kargar Street, 1439957131 Tehran, Iran}

\begin{abstract}
	Feature extraction from financial data is one of the most important problems in market prediction domain for which many approaches have been suggested. Among other modern tools, convolutional neural networks (CNN) have recently been applied for automatic feature selection and market prediction. However, in experiments reported so far, less attention has been paid to the correlation among different markets as a possible source of information for extracting features. In this paper, we suggest a CNN-based framework with specially designed CNNs, that can be applied on a collection of data from a variety of sources, including different markets, in order to extract features for predicting the future of those markets. The suggested framework has been applied for predicting the next day’s direction of movement for the indices of S\&P 500, NASDAQ, DJI, NYSE, and RUSSELL markets based on various sets of initial features. The evaluations show a significant improvement in prediction’s performance compared to the state of the art baseline algorithms.
\end{abstract}

\begin{keyword}
Stock markets prediction\sep Deep learning\sep Convolutional neural networks \sep CNN \sep Feature extraction
\end{keyword}

\end{frontmatter}

%\linenumbers

\section{Introduction}
Financial markets are considered as the heart of the world’s economy in which billions of dollars are traded every day. Clearly, a good prediction of future behavior of markets would be extremely valuable for the traders. However, due to the dynamic and noisy behavior of those markets, making such a prediction is also a very challenging task that has been the subject of research for many years.  In addition to the stock market index prediction, forecasting the exchange rate of currencies, price of commodities and cryptocurrencies like bitcoin are examples of prediction problems in this domain {\citep{shah2014bayesian,zhao2017deep,nassirtoussi2015text,lee2017relative}}. 

Existing approaches for financial market analysis fall into two main groups of fundamental analysis and technical analysis. In technical analysis, historical data of the target market and some other technical indicators are regarded as important factors for prediction. According to the efficient market hypothesis, the price of stocks reflects all the information about them {\citep{fama1970efficient}} while technical analysts believe that prediction of future behavior of the prices in a market is possible by analyzing the previous price data. On the other hand, fundamental analysts examine securities intrinsic value for investment. They look at balance sheets, income statements, cash flow statements and so on to gain insight into future of a company. 

In addition to financial market experts, machine learning techniques have proved to be useful for making such predictions. Artificial neural networks and support vector machine are the most common algorithms that have been utilized for this purpose {\citep{guresen2011using, kara2011predicting, wang2015forecasting}}. Statistical methods, random forests {\citep{khaidem2016predicting}}, linear discriminant analysis, quadratic discriminant analysis, logistic regression and evolutionary computing algorithms, especially genetic algorithm, {\citep{hu2015application, brown2013dynamic, hu2015stock, atsalakis2009surveying}} are among other tools and techniques that have been applied for feature extraction from raw financial data and/or making predictions based on a set of features {\citep{ou2009prediction,ballings2015evaluating}. 
	
Deep learning (DL) is a class of modern tools that is suitable for automatic features extraction and prediction {\citep{lecun2015deep}}. In many domains, such as machine vision and natural language processing, DL methods have been shown to be able to gradually construct useful complex features from raw data or simpler features {\citep{he2016deep,lecun2015deep}}.
Since the behavior of stock markets is complex, nonlinear and noisy, it seems that extracting features that are informative enough for making predictions is a core challenge, and DL seems to be a promising approach to that. Algorithms like Deep Multilayer Perceptron (MLP) {\citep{yong2017stock}}, Restricted Boltzmann Machine (RBM) {\citep{cai2012feature, zhu2014stock}}, Long Short-Term Memory (LSTM) {\citep{chen2015lstm, fischer2018deep}}, Auto-Encoder (AE) {\citep{bao2017deep}} and Convolutional Neural Network (CNN) {\citep{gunduz2017intraday,di2016artificial} are famous deep learning algorithms utilized to predict stock markets. 

It is important to pay attention to the diversity of the features that can be used for making predictions. The raw price data, technical indicators which come out of historical data, other markets with connection to the target market, exchange rates of currencies, oil price and many other information sources can be useful for a market prediction task. Unfortunately, it is usually not a straightforward task to aggregate such a diverse set of information in a way that an automatic market prediction algorithm can use them. So, most of the existing works in this field have limited themselves to a set of technical indicators representing a single market’s recent history {\citep{kim2003financial, zhang2009stock}}.

Another important subject in the field is automatic feature extraction. Since the initial features are defined to be used by human experts, they are simple and even if they were chosen by a finance expert who has enough knowledge and experience in this domain, they may not be the best possible choices for making predictions by machines. In other words, an automatic approach to stock market prediction ideally is one that can extract useful features from different sources of information that seem beneficial for market prediction, train a prediction model based on those extracted features and finally make predictions using the resulted model.  The focus of this paper is on the first phase of this process, that is to design a model for extracting features from several data sources that contain information from historical records of relevant markets. This data includes initial basic features such as raw historical prices, technical indicators or fluctuation of those features in the past days. Regarding the diversity of the input space and possible complexity of the feature space that maybe required for a good prediction, a deep learning algorithm like CNN seems to be a promising approach for such a feature extraction problem. 

To the best of our knowledge, convolutional neural networks, CNN, has been applied in a few studies for stock market prediction {\citep{gunduz2017intraday,di2016artificial}. Periso \& Honchar {\citep{di2016artificial}} used a CNN which took a one-dimensional input for making prediction only based on the history of closing prices while ignoring other possible sources of information like technical indicators. Gunduz et al. {\citep{gunduz2017intraday}} took advantage of a CNN which was capable of using technical indicators as well for each sample. However, it was unable to consider the correlation which could exist between stock markets as another possible source of information. In addition, structure of used CNN was inspired by previous works in Computer Vision, while there is fundamental difference between Computer Vision and Stock market prediction. Since in stock market prediction features interaction are radically different from pixel’s interaction with each other, using $3\times3$ or $5\times5$ filters in convolutional layer may not be the best option. It seems cleverer to design filters of CNN based on financial facts instead of papers in Computer Vision.

We develop our framework based on CNN due to its proven capabilities in other domains as well as mentioned successful past experiments reported in market prediction domain. As a test case, we will show how CNN can be applied in our suggested framework, that we call CNNpred, to capture the possible correlations among different sources of information for extracting combined features from a diverse set of input data from five major U.S. stock market indices: S\&P 500, NASDAQ, Dow Jones Industrial Average, NYSE and RUSSELL, as well as other sources of information including economic data, exchange rate of currencies, future contracts, price of commodities, important indices of markets around the world and price of major companies in U.S. market. Furthermore, the filters are designed in a way that is compatible with financial characteristic of features.
 
The main contributions of this work can be summarized as follows:
\begin{itemize}
	\item Aggregating several sources of information in a CNN-based framework for feature extraction and market prediction. Since financial markets behavior is affected by many factors, it is important to gather related information as much as possible. Our initial feature set covers different aspects of stock related sources of data pretty well and basically, it can be easily extended to cover other possible sources of information.
	\item To our knowledge, this is the first work suggesting a CNN which takes a 3-dimensional tensor aggregating and aligning a diverse set of features as input and is trained to extract features useful for predicting each of the pertinent stock markets afterward.
\end{itemize}

The rest of this paper is organized as follows: In section 2, related works and researches are presented. Then, in section 3, we introduce a brief background on related techniques in the domain.  In section 4, the proposed method is presented in details followed by introduction of various utilized features in section 5. Our experimental setting and results are reported in section 6. In section 7 we discuss the results and there is a conclusion in section 8. 

\section{Related works}
Different methods in stock prediction domain can be categorized into two groups. The first class includes algorithms try to improve the performance of prediction by enhancing the prediction models, while the second class of algorithms focuses on improving the features based on which the prediction is made. 

In the first class of the algorithms that focus on the prediction models, a variety of tools have been used, including Artificial Neural Networks (ANN), naive Bayes, SVM and random forests. The most popular tool for financial prediction seems to be ANN {\citep{krollner2010financial}}. In {\citep{kara2011predicting}}, a comparison between performance of ANN and SVM were done. Ten technical indicators were passed to these two classifiers in order to forecast directional movement of the Istanbul Stock Exchange (ISE) National 100 Index. Authors found that ANN’s ability in prediction is significantly better than SVM. 

Feedforward ANNs are popular types of ANNs that are capable of predicting both price movement direction and price value. Usually shallow ANNs are trained by back-propagation algorithm {\citep{hecht1992theory, hagan1994training}}. While obstacles like the noisy behavior of stock markets and complexity of feature space make ANNs’ learning process to converge to suboptimal solutions, sometimes local search algorithms like GA or SA take responsibility of finding initial or final optimal weights for neural networks that are then used for prediction {\citep{kim2000genetic, qiu2016application, qiu2016predicting}}. In {\citep{qiu2016application}, authors used genetic algorithm and simulated annealing to find initial weights of an ANN, and then back-propagation algorithm is used to train the network. This hybrid approach outperformed the standard ANN-based methods in prediction of Nikkei 225 index return. With slight modifications in {\citep{qiu2016predicting}}, genetic algorithm was successfully utilized to find optimized weights of an ANN in which technical indicators were utilized to predict the direction of Nikkei 225 index movement.

Authors of {\citep{zhong2017forecasting}} have applied PCA and two variations of it in order to extract better features. A collection of different features was used as input data while an ANN was used for prediction of S\&P 500. The results showed an improvement of the prediction using the features generated by PCA compared to the other two variations of that. The reported accuracy of predictions varies from 56\% to 59\% for different number of components used in PCA.
Another study on the effect of features on the performance of prediction models has been reported in {\citep{patel2015predicting}}. This research uses common tools including ANN, SVM, random forest and naive Bayes for predicting directional movement of famous indices and stocks in Indian stock market. This research showed that mapping the data from a space of ten technical features to another feature space that represents trends of those features can lead to an improvement in the performance of the prediction. 

According to mentioned researches and similar works, when a shallow model is used for prediction, the quality of features by which the input data is represented has a critical role in the performance of the prediction. The simplicity of shallow models can avoid them from achieving effective mappings from input space to successful predictions. So, with regards to availability of large amounts of data and emerging effective learning methods for training deep models, researchers have recently turned to such approaches for market prediction. An important aspect of deep models is that they are usually able to extract rich sets of features from the raw data and make predictions based on that. So, from this point of view, deep models usually combine both phases of feature extraction and prediction in a single phase. 

Deep ANNs, that are basically neural networks with more than one hidden layers, are among the first deep methods used in the domain. In {\citep{moghaddam2016stock}}, authors predicted NASDAQ prices based on the historical price of four and nine days ago. ANNs with different structures, including both deep and shallow ones, were examined in order to find appropriate number of hidden layers and neurons inside them. The experiments proved the superiority of deep ANNs over shallow ones. In {\citep{arevalo2016high}, authors used a deep ANN with five hidden layers to forecast Apple Inc’s stock price during the financial crisis. For each minute of trading three features were extracted from the fluctuation of price inside that time period. Outputs showed up to about 65\% directional accuracy. In {\citep{yong2017stock}}, an ANN with three hidden layers was utilized to predict the index price of Singapore’s stock market. Historical prices of the last ten days were fed to a deep ANN in order to predict the future price of next one to five days. This experiment reported that the highest performance was achieved for one day ahead prediction with MAPE of 0.75.

In {\citep{chong2017deep}}, authors draw an analogy between different data representation methods including RBM, Auto-encoder and PCA applied on raw data with 380 features. The resulting representations were then fed to a deep ANN for prediction. The results showed that none of the data representation methods has superiority over the others in all of the tested experiments. 

Recurrent Neural Networks are a kind of neural networks that are specially designed to have internal memory that enables them to extract historical features and make predictions based on them. So, they seem fit for the domains like market prediction in which historical behavior of markets has an important role in prediction. LSTM is one of the most popular kinds of RNNs. In {\citep{nelson2017stock}}, technical indicators were fed to an LSTM in order to predict the direction of stock prices in the Brazilian stock market. According to the reported results, LSTM outperformed MLP, by achieving an accuracy of 55.9\%.

Convolutional Neural Network is another deep learning algorithm applied in stock market prediction after MLP and LSTM while its ability to extract efficient features has been proven in many other domains as well. In {\citep{di2016artificial}}, CNN, LSTM and MLP were applied to the historical data of close prices of S\&P 500 index. Results showed that CNN outperformed LSTM and MLP with accuracy of 53.6\% while LSTM and MLP had accuracy of 52.2\% and 52.1\% respectively. 

Based on some reported experiments, the way the input data is designed to be fed and processed by CNN has an important role in the quality of the extracted feature set and the final prediction. For example, CNN was used in {\citep{gunduz2017intraday}} in which data of 10 days of 100 companies in Borsa Istanbul were utilized to produce technical indicators and time-lagged features. Then, a CNN was applied to improve the feature set. The reported comparison between CNN and logistic regression shows almost no difference between two methods. In another attempt to improve the prediction, features were clustered into different groups and similar features were put beside each other. The experiments showed that this preprocessing step has improved the performance of CNN to achieve F-measure of 56\%.

Table \ref{table:summary} summarizes explained papers in terms of initial feature set, feature extraction algorithm and prediction method. As it can be seen there is a tendency toward deep learning models in recent publications, due to the capability of these algorithms in automatic feature extraction from raw data. However, most of the researchers have used only technical indicators or historical price data of one market for prediction while there are various sources of data which could enhance accuracy of prediction of stock market. In this paper, we are going to introduce a novel CNN-based framework that is designed to aggregate several sources of information in order to automatically extract features to predict direction of stock markets.

\begin{table}
	\begin{center}
		%\tiny
		\resizebox{1\textwidth}{!}
		{
			\begin{tabular}{c c c c c }
				\hline
				Author/year & Target Data & Feature Set & \specialcell{Feature \\Extraction}  &  \specialcell{Prediction\\ Method}  \\ 
				\hline \hline
				{\citep{kara2011predicting}} & \specialcell{Borsa Istanbul \\BIST 100 Index} & technical indicator & \specialcell{ANN} & \specialcell{ANN\\ SVM} \\\\
				{\citep{patel2015predicting}} & \specialcell{4 Indian stocks\\ \& indices} & technical indicator & \specialcell{ANN} & \specialcell{ANN-SVM\\ RF-NB} \\\\
				{\citep{qiu2016application}} & \specialcell{Nikkei 225 \\index} & \specialcell{financial indicator \\macroeconomic data} & ANN & \specialcell{GA+ANN \\SA+ANN} \\\\
				{\citep{qiu2016predicting}}& \specialcell{Nikkei 225 \\index} & technical indicator & ANN & GA+ANN \\\\
				{\citep{nelson2017stock}}& \specialcell{Brazil Bovespa\\5 stocks} & technical indicator & LSTM & LSTM \\\\
				{\citep{di2016artificial}}& \specialcell{S\&P 500 index} & price data & \specialcell{MLP-RNN-CNN \\wavelet+CNN} & \specialcell{MLP\\RNN\\CNN} \\\\
				{\citep{moghaddam2016stock}}& \specialcell{NASDAQ } & price data  & ANN-DNN & ANN-DNN  \\\\
				{\citep{arevalo2016high}}& \specialcell{AAPL Inc. } & 3 extracted features  & DNN & DNN \\\\
				{\citep{zhong2017forecasting}}& \specialcell{S\&P 500 index } & \specialcell{various sources \\ of data}  & PCA & ANN  \\\\
				{\citep{yong2017stock}}& \specialcell{Singapore STI } & price data & DNN & DNN  \\\\
				{\citep{chong2017deep}}& \specialcell{Korea KOSPI \\38 stock returns} & price data & \specialcell{PCA-RBM \\ AE} & DNN \\\\
				{\citep{gunduz2017intraday}}& \specialcell{Borsa Istanbul \\BIST 100 stocks} & \specialcell{technical indicator \\temporal feature} & \specialcell{Clustering\\CNN} & CNN \\\\
				Our method& \specialcell{U.S. 5 \\major indices} & \specialcell{various sources \\of data} & \specialcell{3D representation\\ of data+CNN} & CNN \\\\
			
			\end{tabular}
			}
		\caption{Summary of explained papers}
		\label{table:summary}
	\end{center}
\end{table}

\section{Background}
Before presenting our suggested approach, in this section, we review the convolutional neural network that is the main element of our framework.

\subsection{Convolutional Neural Network}
LeCun and his colleagues introduced convolutional neural networks in 1995 \citep{lecun1995convolutional,gardner1998artificial}. CNN has many layers which could be categorized into input layer, convolutional layers, pooling layers, fully connected layers and output layer. 

\subsubsection{Convolutional layer}
The convolutional layer is supposed to do the convolution operation on the data. In fact, input could be considered as a function, filter applied to that is another function and convolution operation is an algorithm used to measure changes caused by applying filter on the input. Size of a filter shows the coverage of that filter. Each filter utilizes a shared set of weights to perform the convolutional operation. Weights are updated during the process of training. 

Let’s posit input of layer $l-1$ is an $N\times N$ matrix and $F\times F$ convolutional filters are used. Then, input of layer $l$ is calculated according to Eq \ref{eq : conv}. Fig \ref{fig: conv} shows applying filter to the input data in order to get value of $v_{1,1}$ in the next layer. Usually, output of each filter is passed through an activation function before entering the next layer. Relu (Eq \ref{eq:relu}) is a commonly used nonlinear activation function.

\begin{equation}
	v_{i,j}^{l} = \delta(\sum_{k=0}^{F-1} \sum_{m=0}^{F-1} w_{k,m} V_{i+k,j+m}^{l-1})
	\label{eq : conv}
\end{equation}

\begin{figure}[!h]
	\makebox[\textwidth]{\includegraphics[width=0.7\paperwidth]{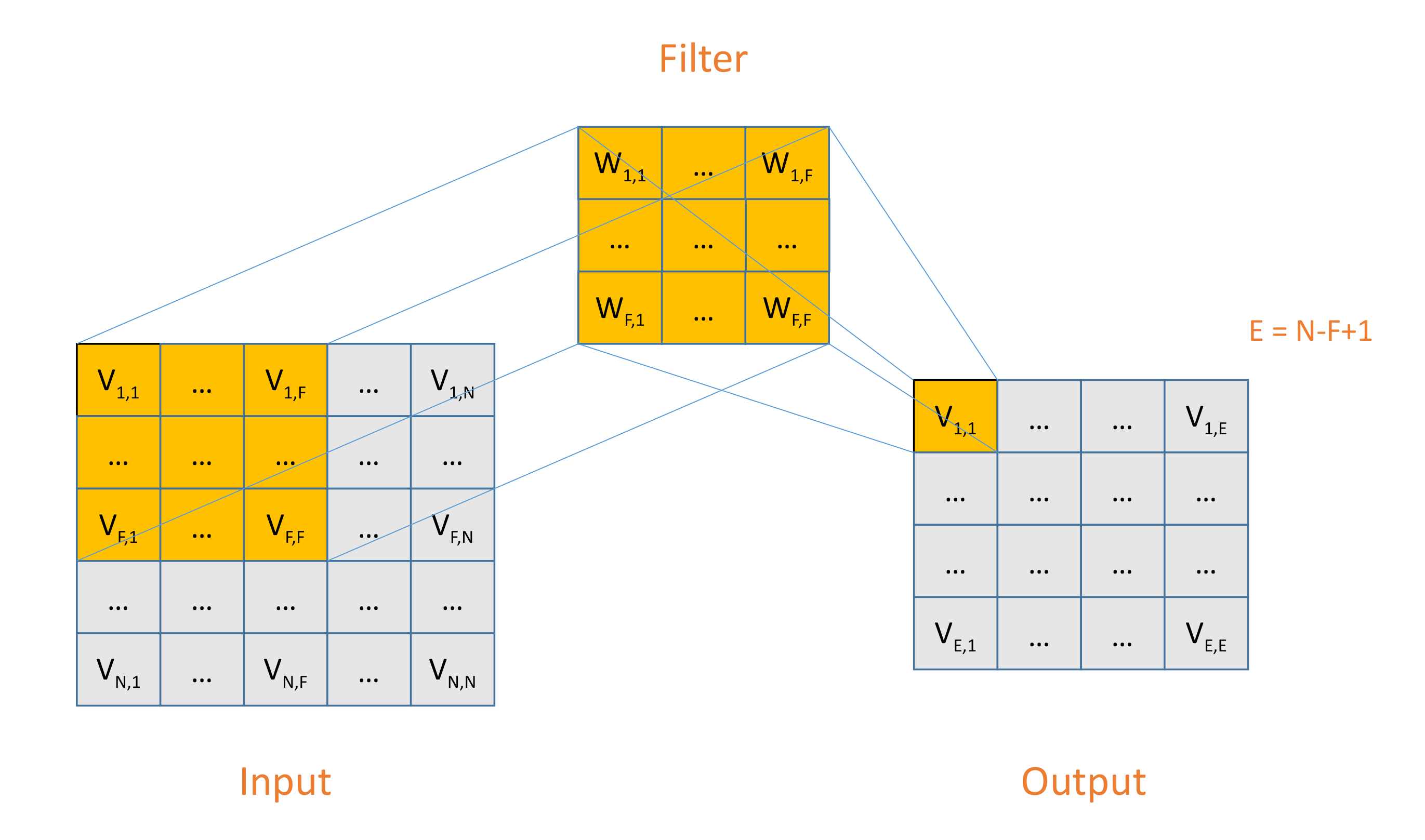}}
	\caption{Applying filter($F\times F$) to the input data($N\times N$) in order to get value of $V_{1,1}$ in the next layer}
	\label{fig: conv}
\end{figure}

In the Eq \ref{eq : conv}, $v^{l}_{i,j}$ is the value at row $i$, column $j$ of layer $l$, $w_{k,m}$ is the weight at row $k$, column $m$ of filter and $\delta$ is the activation function.  

\begin{equation}
	f(x) = max(0,x)
	\label{eq:relu}
\end{equation}

\subsubsection{Pooling layer}
Pooling layer is responsible for subsampling the data. This operation, not only reduces the computational cost of the learning process, but also it is a way for handling the overfitting problem in CNN. Overfitting is a situation that arises when a trained model makes too fit to the training data, such that it cannot generalize to the future unseen data. It has a connection to the number of parameters that are learned and the amount of data that the prediction model is learned from. Deep models, including CNNs, usually have many parameters so they are prone to overfitting more than shallow models. Some methods have been suggested to avoid overfitting. Using pooling layers in CNNs can help to reduce the risk of overfitting. All the values inside a pooling window are converted to only one value. This transformation reduces the size of the input of the following layers, and hence, reduces the number of the parameters that must be learned by the model, that in turn, lowers the risk of overfitting. Max pooling is the most common type of pooling in which the maximum value in a certain window is chosen.

\subsubsection{Fully connected layer}
At the final layer of a CNN, there is an MLP network which is called its fully connected layer. It is responsible for converting extracted features in the previous layers to the final output. The relation between two successive layers is defined by Eq \ref{eq:activation}

\begin{equation}
v_{i}^{j} = \delta(\sum_k v_{k}^{j-1}w_{k,i}^{j-1})
\label{eq:activation}
\end{equation}

In Eq \ref{eq:activation}, $v_{i}^{j}$ is the value of neuron $i$ at the layer $j$, $\delta$ is activation function and weight of connection between neuron $k$ from layer $j-1$ and neuron $i$ from layer $j$ are shown by $w_{k,i}^{j-1}$.

\subsection{Dropout}
In addition to pooling, we have also used another technique called dropout that was first developed for training deep neural networks. The idea behind the dropout technique is to avoid the model from learning too much from the training data. So, in each learning cycle during the training phase, each neuron has a chance equal to some \textit{dropout rate}, to not be trained in that cycle. This avoids the model from being too flexible, and so, helps the learning algorithm to converge to a model that is not too much fit to the training data, and instead, can be generalized well for prediction the unlabeled future data \citep{hinton2012improving,srivastava2014dropout}.

\section{Proposed CNN: CNNpred}
CNN has many parameters including the number of layers, number of filters in each layer, dropout rate, size of filters in each layer, initial representation of input data and so on which should be chosen wisely to get the desired outcomes. Although $3\times 3$ and $5\times 5$ filters are quite common in image processing domain, we think that size of each filter should be determined according to financial interpretation of features and their characteristics rather than just following previous works in image processing. Here we introduce the architecture of CNNPred, a general CNN-based framework for stock market prediction. CNNPred has two variations that are referred to as 2D-CNNpred and 3D-CNNpred. We explain the framework in four major steps:  representation of input data, daily feature extraction, durational feature extraction and final prediction.

Representation of input data: CNNpred takes information from different markets and uses it to predict the future of those markets. As we mentioned 2D-CNNpred and 3D-CNNpred take different approaches for constructing prediction models. The goal of the first approach is to find a general model for mapping the history of a market to its future fluctuations and by "general model” we mean a model that is valid for several markets. In other words, we assume that the true mapping function from the history to the future is the one that is correct for many markets. For this goal, we need to design a single model that is able to predict the future of a market based on its own history, however, to extract the desired mapping function, that model needs to be trained by samples from different markets. 2D-CNNpred follows this general approach, but in addition to modeling the history of a market as the input data, it also uses a variety of other sources of information as well. In 2D-CNN-pred all this information is aggregated and fed to a specially designed CNN as a two-dimensional tensor, and that’s why it is called 2D-CNNpred. 
On the other hand, the second approach, 3D-CNNpred, assumes that different models are needed for making predictions in different markets, but each prediction model can use information from the history of many markets. In other words, 3D-CNNpred, unlike 2D-CNNpred, does not train a single prediction model that can predict the future of each market given its own historical data, but instead, it extracts features from the historical information of many markets and uses them to train a separate prediction model for each market. The intuition behind this approach is that the mechanisms that dictate the future behavior of each market differs, at least slightly, from other markets. However, what happens in the future in a market, may depend on what happens inside and outside that certain market. Based on this intuition, 3D-CNNpred uses a tensor with three dimensions, to aggregate historical information from various markets and feed it to a specially designed CNN to train a prediction model for each market. Although the structure of the model is the same for all the markets, the data that is used for training is different for each market. In other words, in 3D-CNNpred, each prediction model can see all the available information as input, but is trained to predict the future of a certain market based on that input. One can expect that 3D-CNNpred, unlike 2D-CNNpred, will be able to combine information from different markets into high-level features before making predictions. Fig \ref{fig: twomodel} shows a schema of how data is represented and used in CNNpred’s variations.

\begin{figure}[!h]
	\makebox[\textwidth]{\includegraphics[width=0.7\paperwidth]{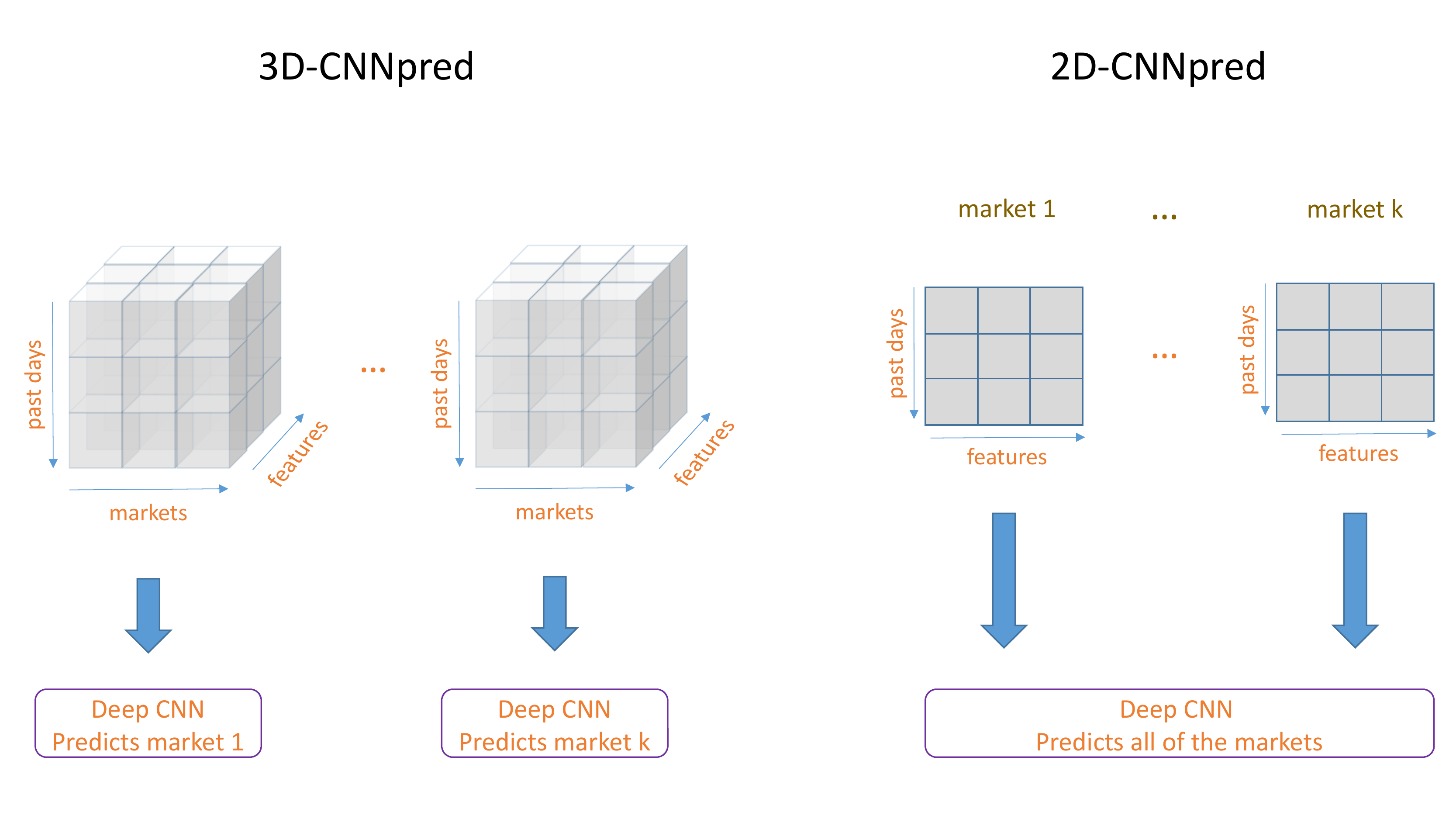}}
	\caption{The structure of input data in two variations of CNNpred}
	\label{fig: twomodel}
\end{figure}

Daily feature extraction: Each day in the historical data is represented by a series of features like opening and closing prices. The traditional approach to market prediction is to analyze these features for example in the form of candlesticks, probably by constructing higher level features based on them, in order to predict the future behavior of the market. The idea behind the design of first layer of CNNpred comes from this observation. In the first step of both variations of CNNpred, there is a convolutional layer whose task is to combine the daily features into higher level features for representing each single day of the history.

Durational feature extraction: Some other useful information for predicting the future behavior of a market comes from studying the behavior of the market over time. Such a study can give us information about the trends that appear in the market’s behavior, and find patterns that can predict the future based on them. So it is important to combine features of consecutive days of data to gather high-level features representing trends or reflecting the market’s behavior in certain time intervals. Both 2D-CNNpred and 3D-CNNpred data have layers that are supposed to combine extracted features in the first layer and produce even more sophisticated features summarizing the data in some certain time interval. 

Final prediction: At the final step, the features that are generated in previous layers are converted to a one-dimensional vector using a flattening operation and this vector is fed to a fully connected layer that maps the features to a prediction.  

In the next two sections, we will explain the general design of 2D-CNNpred and 3D-CNNpred as well as how they have been adopted for the data set that we have used in the specific experiments performed in this paper. In our experiments, we have used data from 5 different indices. Each index has 82 features that means each day of the history of a market is represented by 82 features. The 82 gathered features are selected in a way that form a complete feature set and consist of economic data, technical indicators, big U.S. companies, commodities, exchange rate of currencies, future contracts and world’s stock indices. The length of the history is 60 days that is for each prediction, the model can use information from 60 last days.

\subsection{2D-CNNpred}
Representation of input data: As we mentioned before, the input to the 2D-CNNpred is a two-dimensional matrix. The size of the matrix depends on the number of features that represent each day, as well as the number of days back into the history, that is used for making a prediction. If the input used for prediction consists of $d$ days each represented by $f$ features then the size of input tensor will be $d\times f$. 

Daily feature extraction: To extract daily features in 2D-CNNpred, $1\times$\textit{number of initial features} filters are utilized. Each of those filters covers all the daily features and can combine them into a single higher level feature, so using this layer, 2D-CNNpred can construct different combinations of primary features. It is also possible for the network to drop useless features by setting their corresponding weights in filters equal to zero. So this layer works as an initial feature extraction/feature selection module. Fig \ref{fig: 1by-feature-filter} represents application of a simple filter on the input data.

\begin{figure}[!h]
	\makebox[\textwidth]{\includegraphics[width=0.6\paperwidth]{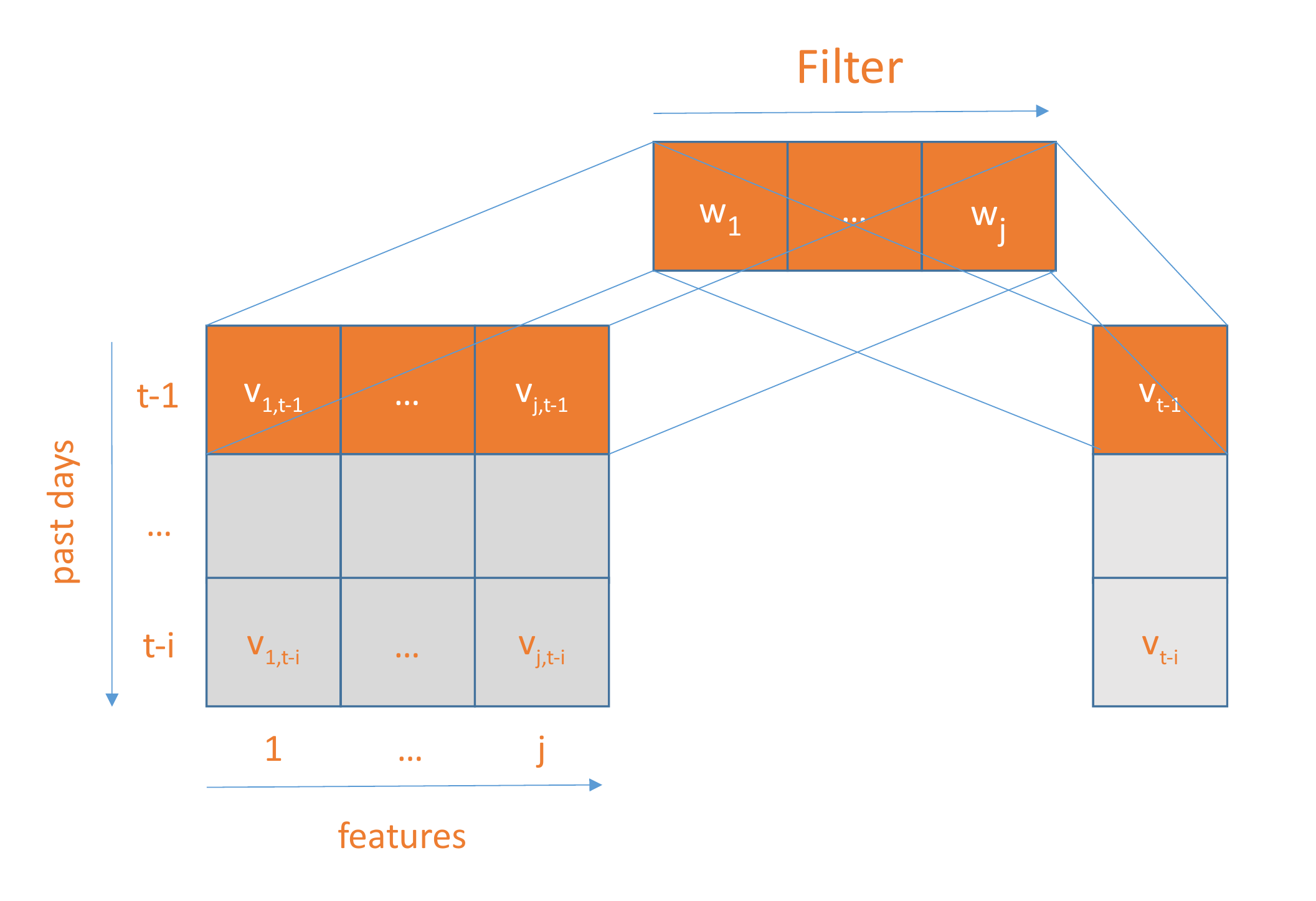}}
	\caption{Applying a $1\times$\textit{number of features} filter to 2D input tensor.}
	\label{fig: 1by-feature-filter}
\end{figure}

Durational feature extraction: While the first layer of 2D-CNNpred extracts features out of primary daily features, the following layers combine extracted features of different days to construct higher level features for aggregating the available information in certain durations. Like the first layer, these succeeding layers use filters for combining lower level features from their input to higher level ones. 2D-CNNpred uses $3\times 1$ filters in the second layer. Each of those filters covers three consecutive days, a setting that is inspired by the observation that most of the famous candlestick patterns like Three Line Strike and Three Black Crows, try to find meaningful patterns in three consecutive days {\citep{nison1994beyond,bulkowski2012encyclopedia, achelis2001technical}. We take this as a sign of the potentially useful information that can be extracted from a time window of three consecutive times unites in the historical data. The third layer is a pooling layer that performs a $2\times 1$ max pooling, that is a very common setting for the pooling layers. After this pooling layer and in order to aggregate the information in longer time intervals and construct even more complex features, 2D-CNNpred uses another convolutional layer with $3\times 1$ filters followed by a second pooling layer just like the first one.

Final prediction: Produced features generated by the last pooling layer are flattened into a final feature vector. This feature vector is then converted to a final prediction through a fully connected layer. Sigmoid (Eq \ref{eq: sigmoid}) is the activation function that we choose for this layer. Since the output of sigmoid is a number in [0-1] interval, the prediction that is made by 2D-CNNpred for a market can be interpreted as the probability of an increase in the price of that market for the next day, that is a valuable piece of information. Clearly, it is rational to put more money on a stock that has a higher probability of going up. On the other hand, stocks with a low probability of going up are good candidates for short selling. However, in our experiments, we discretize the output to either 0 or 1, whichever is closer to the prediction. 

\begin{equation}
	f(x) = \frac{1}{1+\exp(x)}
	\label{eq: sigmoid}
\end{equation}

A sample configuration of 2D-CNNpred: As we mentioned before the input we used for each prediction consists of 60 days each represented by 82 features. So, the input to the 2D-CNNpred is a matrix of 60 by 82. The first convolutional layer uses eight $1\times 82$ filters after which there are two convolutional layers with eight $3\times 1$ filters, each followed by a layer of $2\times 1$ max-pooling. The final flattened feature vector contains 104 features that are fed to the fully connected layer to produce the final output. Fig \ref{fig: 2D-CNNpred} shows a graphical visualization of described process.

\begin{figure}[!h]
	\makebox[\textwidth]{\includegraphics[width=0.8\paperwidth]{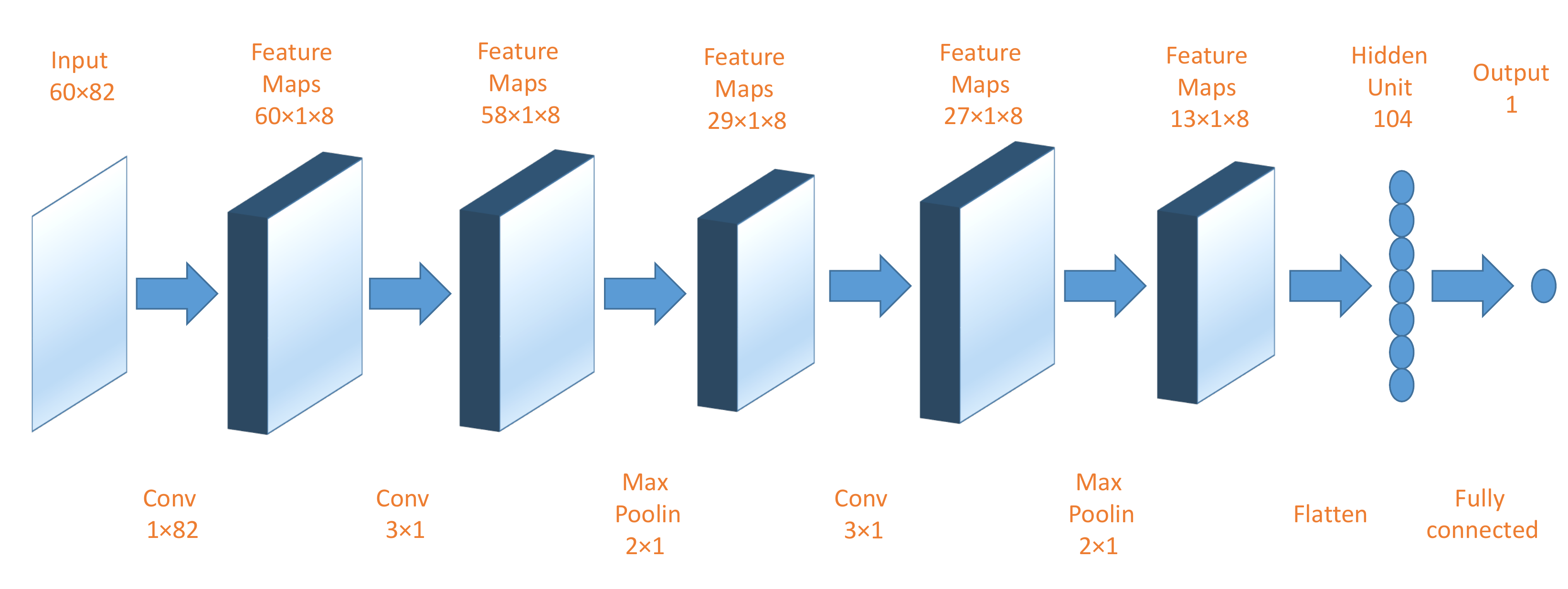}}
	\caption{Graphical Visualization of 2D-CNNpred}
	\label{fig: 2D-CNNpred}
\end{figure}

\subsection{3D-CNNpred}
Representation of input data: 3D-CNNpred, unlike 2D-CNNpred, uses a three-dimensional tensor to represent data. The reason is that each sample that is fed to 3D-CNNpred, contains information from several markets. So, the initial daily features, the days of the historical record and the markets from which the data is gathered form the three dimensions of the input tensor. Suppose our dataset consists $i$ different markets, $k$ features for each of these markets and our goal is to predict day $t$ based on past $j$ days. Fig \ref{fig: data-rep-3d} shows how one sample of the data would be represented.

\begin{figure}[!h]
	\makebox[\textwidth]{\includegraphics[width=0.4\paperwidth]{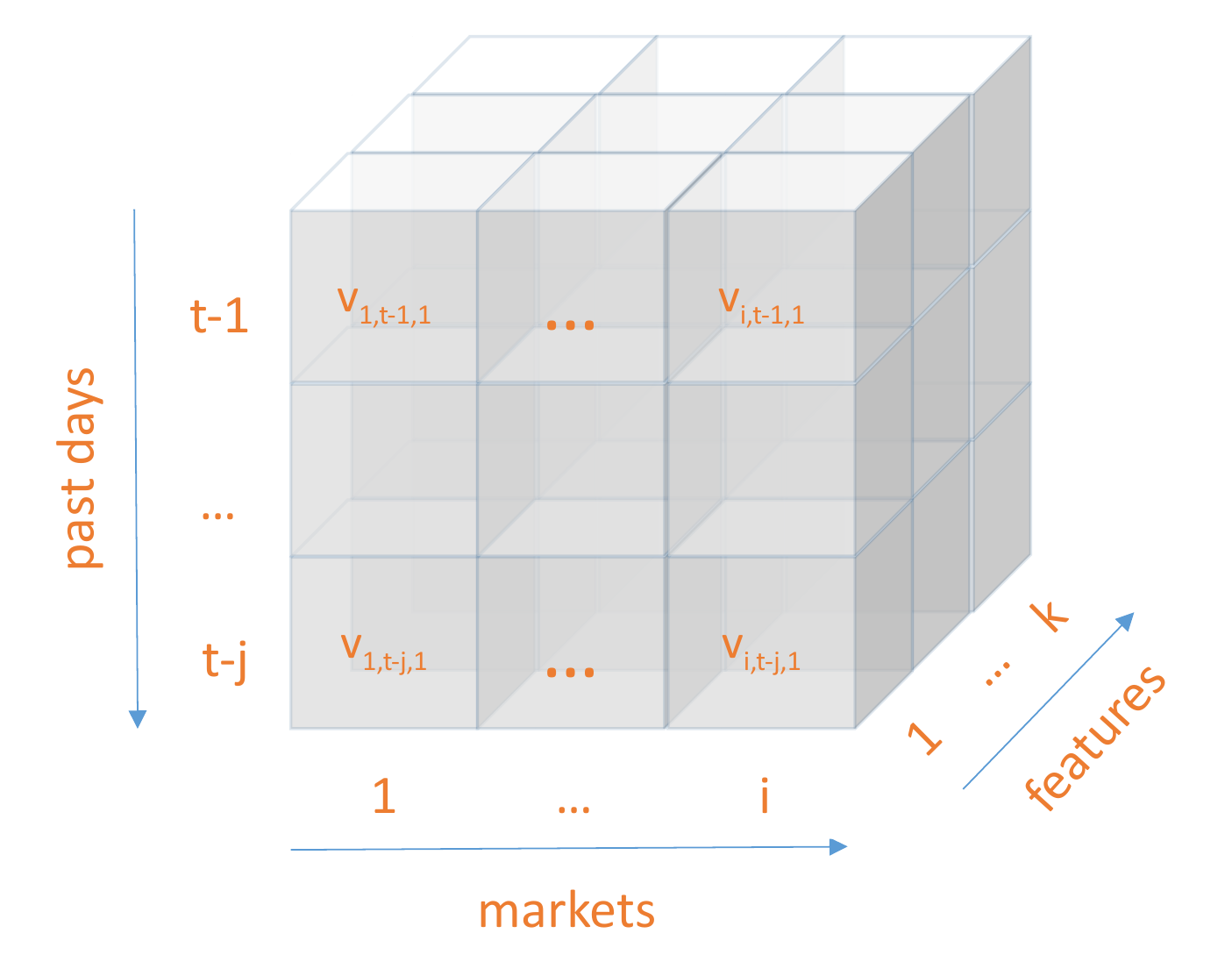}}
	\caption{Representation of input data in 3D-CNNpred based on $k$ primary features, $i$ related markets and $j$ days before the day of prediction}
	\label{fig: data-rep-3d}
\end{figure}

Daily feature extraction: The first layer of filters in 3D-CNNpred is defined as a set of $1\times 1$ convolutional filters, while the primary features are represented along the depth of the tensor. Fig \ref{fig: 1by1-filter} shows how a $1\times 1$ filter works. This layer of filters is responsible for combining subsets of basic features that are available through the depth of the input tensor into a set of higher level features. The input tensor is transformed by this layer into another tensor whose width and height is the same but its depth is equal to the number of $1\times 1$ convolutional filters of layer one. Same as 2D-CNNpred, the network has the capability to act as a feature selection/extraction algorithm.

\begin{figure}[!h]
	\makebox[\textwidth]{\includegraphics[width=0.7\paperwidth]{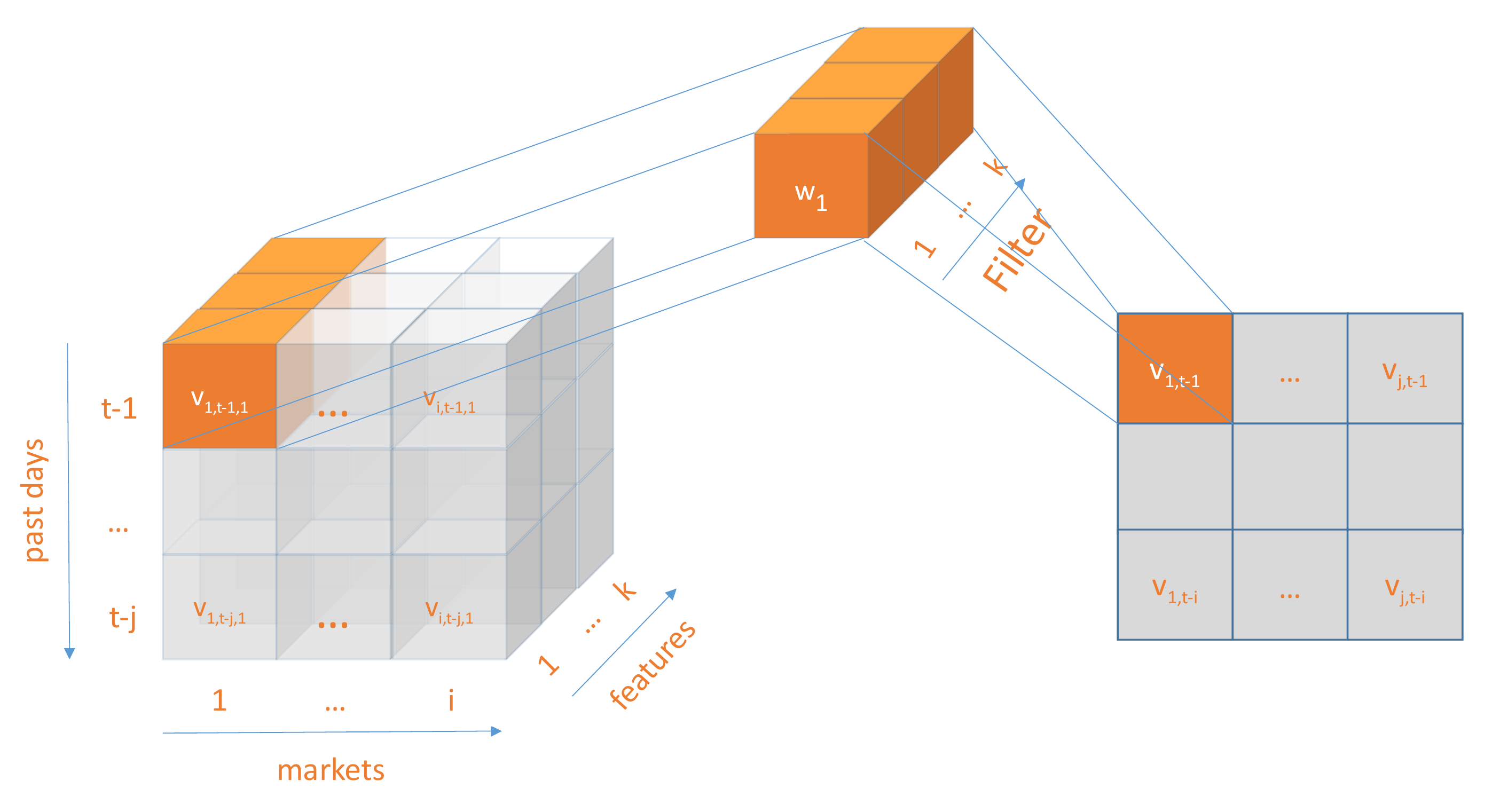}}
	\caption{Applying a $1\times1$ filter to the first part of the 3D input tensor.}
	\label{fig: 1by1-filter}
\end{figure}

Durational feature extraction: In addition to daily features, 3D-CNNpred’s input data provides information about other markets. Like 2D-CNNpred, the next four layers are dedicated to extracting higher level features that summarize the fluctuation patterns of the data in time. However, in 3D-CNNpred, this is done over a series of markets instead of one. So, the width of the filters in the second convolutional layer is defined in a way that covers all the pertinent markets. Same as 2D-CNNpred and motivated by the same mentioned reason, the height of filters is selected to be 3 so as to cover three consecutive time units. Using this setting, the size of filters in the second convolutional layer is $3\times$\textit{number of markets}. The next three layers, like those of 2D-CNNpred, are defined as a $2\times 1$ max pooling layer, another $3\times 1$ convolutional layer followed by a final $2\times 1$ max pooling layer. 

Final prediction: Same as 2D-CNNpred, here in 3D-CNNpred the output of the durational feature extraction phase is flattened and used to produce the final results.

A sample configuration of 3D-CNNpred: In our experiments, the input to the 3D-CNNpred is a matrix of 60 by 5 with depth of 82. The first convolutional layer uses eight filters to perform $1\times 1$ convolutional operation, after which there is one convolutional layer with eight $3\times 5$ filters followed by $2\times 1$ max pooling layer. Then, another convolutional layer utilizes eight $3\times 1$ filters, again followed by a $2\times 1$ max-pooling layer generate the final 104 features. In the end, a fully connected layer converts 104 neurons to 1 neuron and produces the final output. Fig \ref{fig: 3D-CNNpred} shows a graphical visualization of the process. 

\begin{figure}[!h]
	\makebox[\textwidth]{\includegraphics[width=0.8\paperwidth]{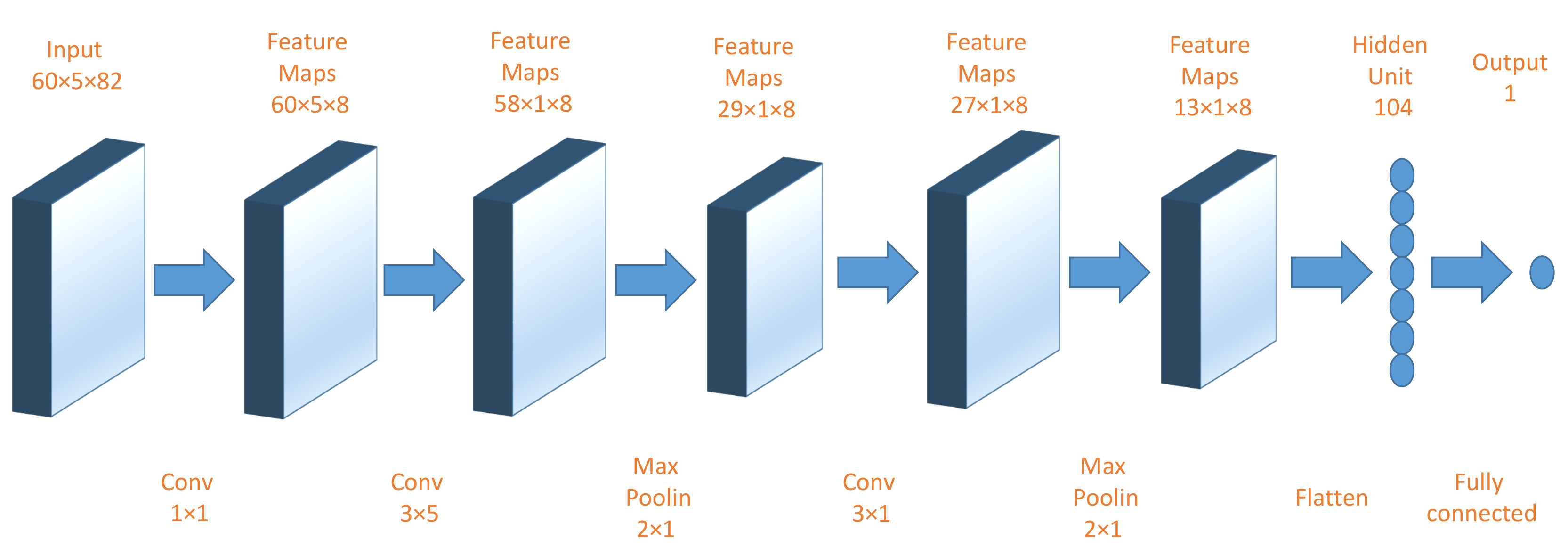}}
	\caption{Graphical Visualization of 3D-CNNpred}
	\label{fig: 3D-CNNpred}
\end{figure}

\section{Initial feature set for each market}
As we mentioned before, our goal is to develop a model for prediction of the direction of movements of stock market prices or indices. We applied our approach to predict the movement of indices of S\&P 500, NASDAQ, Dow Jones Industrial Average, NYSE and RUSSELL market. For this prediction task, we use 82 features for representing each day of each market. Some of these features are market-specific while the rest are general economic features and are replicated for every market in the data set. This rich set of features could be categorized in eight different groups that are primitive features, technical indicators, economic data, world stock market indices, the exchange rate of U.S. dollar to the other currencies, commodities, data from big companies of U.S. market and future contracts. We briefly explain different groups of our feature set here and more details about them can be found in Appendix I.

\begin{itemize}
	\item Primitive features: Close price and which day of week prediction is supposed to happen are primitive features used in this work. 
	\item Technical indicators: Technical analysts use technical indicators which come out of historical data of stocks like price, volume and so on to analyze short-term movement of prices. They are quite common in stock market research. The moving averages are examples of this type of features.
	\item Economic data: Economic data reflects whether the economy of a country is doing well or not. In addition to the other effective factors, investors usually take a look at these indicators so as to gain insight into future of stock market. Information coming from Treasury bill belongs to this category.
	\item World stock markets: Usually, stock markets all over the world have interaction with each other because of the phenomenon of globalization of economy. This connection would be more appreciated when we consider time difference in various countries which makes it possible to gain information about future of a country\textsc{\char13}s market by monitoring other countries markets. For instance, effect of other countries stock market like China, Japan and South Korea on U.S. market.
	\item The exchange rate of U.S. dollar: There are companies that import their needs from other countries or export their product to other countries. In these cases, value of U.S. dollar to other currencies like Canadian dollar and European Euro make an important role in the fluctuation of stock prices and by extent, the whole market.
	\item Commodities: Another source of information that affects stock market is price of commodities like gold, silver, oil and so on. This kind of information can reflect a view of the global market. This means that the information about the prices of commodities can be useful in prediction of the fluctuations of stock prices.
	\item Big U.S. Companies: Stock market indices are calculated based on different stocks. Each stock carries a weight in this calculation that matches its share in the market. In another word, big companies are more important than small ones in prediction of stock market indices. Examples of that could be Exxon Mobil Corporation and Apple Inc.
	\item Futures contracts: Futures contracts are contracts in which one side of agreement is supposed to deliver stock, commodities and so on in the future. These contracts show expected value of the merchandise in the future. Investors tend to buy stocks that have higher expected value than their current value.  For instance, S\&P 500 Futures, DJI Futures and NASDAQ Futures prices could affect current price of S\&P 500 and other indices.
\end{itemize}
 
\section{Experimental settings and results}
In this section, we describe the settings that are used to evaluate the models, including datasets, parameters of the networks, evaluation methodology and baseline algorithms. Then, the evaluation results are reported.

\subsection{Data gathering and preparation}
The datasets used in this work include daily direction of close of S\&P 500 index, NASDAQ Composite, Dow Jones Industrial Average, NYSE Composite and RUSSELL 2000. Table \ref{table: dataset} shows more information about them. Each sample has 82 features that already have been explained and its assigned label is determined according to the Eq \ref{eq: target}. It is worth mentioning that for each index only technical indicators and primitive features are unique and the other features, like big U.S. companies or price of commodities, are common between different indices.

\begin{table}
	\centering
	\begin{tabular}{c| c }
		Name & Description  \\ 
		\hline \hline
		S\&P 500 & Index of 505 companies exist in S\&P stock market \\
		Dow Jones Industrial Average & Index of 30 major U.S. companies \\  
		NASDAQ Composite & Index of common companies exist in NASDAQ stock market \\
		NYSE Composite & Index of common companies exist in New York Stock Exchange\\
		RUSSEL 2000 &  Index of 2000 small companies in U.S. \\	
	\end{tabular}
	\caption{Description of used indices}
	\label{table: dataset}
\end{table}

\begin{equation}
	target = 
	\begin{cases}
	1 & Close_{t+1}>Close_t \\
	0 & \text{else }\\
	\end{cases}
	\label{eq: target}
\end{equation}

Where $Close_{t}$ refers to the closing price at day t.

This data are from the period of Jan 2010 to Nov 2017. The first 60\% of the data is used for training the models, the next 20\% forms the validation data and the last 20\% is the test data.

Different features could have various ranges. It is usually confusing for learning algorithms to handle features with different ranges. Generally, the goal of data normalization here is to map the values of all features to a single common range, and it usually improves the performance of the prediction model. We use Eq \ref{eq: normalization} for normalizing the input data, where $x_{new}$ is normalized feature vector, $x_{old}$ is the original feature vector, $\bar{x}$ and $\sigma$ are the mean and the standard deviation of original feature.

\begin{equation}
	x_{new} = \frac{x_{old} - \bar{x}}{\sigma}
	\label{eq: normalization}
\end{equation}

\subsection{Evaluation methodology}
Evaluation metrics are needed to compare results of our method with the other methods. Accuracy is one of the common metrics have been used in this area. However, in an imbalanced dataset, it may be biased toward the models that tend to predict the more frequent class. To address this issue, we report the Macro-Averaged-F-Measure that is the mean of F-measures calculated for each of the two classes {\citep{gunduz2017intraday, ozgur2005text}}.

\subsection{Parameters of network}
Numerous deep learning packages and software have been developed. In this work, Keras {\citep{chollet2015keras} was utilized to implement CNN. The activation function of all the layers except the last one is RELU. Complete descriptions of parameters of CNN are listed in Table \ref{table:parameters}.  
\begin{table}
	\centering
	\begin{tabular}{c| c}
		Parameter & Value \\ 
		\hline \hline
		Filter size & \{8, 8, 8\} \\
		Activation function & RELU-Sigmoid \\
		Optimizer & Adam \\
		Dropout rate & 0.1 \\
		Batch size & 128 \\ 
	\end{tabular}
	\caption{Parameters of CNN}
	\label{table:parameters}
\end{table}
  
\subsection{Baseline algorithms}
We compare the performance of the suggested methods with that of the algorithms applied in the following researches. In all the base-line algorithms the same settings reported in the original paper were used.

\begin{itemize}
	\item The first baseline algorithm is the one reported in {\citep{zhong2017forecasting}}. In this algorithm, the initial data is mapped to a new feature space using PCA and then the resulting representation of the data is used for training a shallow ANN for making predictions. 
	\item The second baseline is based on the method suggested in {\citep{kara2011predicting}}, in which the technical indicators reported in Table \ref{table: technical} are used to train a shallow ANN for prediction.
	\item The third baseline algorithm is a CNN with two-dimensional input {\citep{gunduz2017intraday}}. First, the features are clustered and reordered accordingly. The resulting representation of the data is then used by a CNN with a certain structure for prediction. 
	
\end{itemize}

\begin{table}
	\centering
	\begin{tabular}{c| c}
		
		Indicator & Description  \\ 
		\hline \hline
		MA & Simple Moving Average \\
		EMA & Exponential Moving Average \\ 
		MOM & Momentum \\
		\%K & Stochastic \%K \\
		\%D & Stochastic \%D \\
		RSI & Relative Strength Index  \\
		MACD & Moving Average Convergence Divergence \\
		\%R & Larry William’s \%R \\
		A$\setminus$D & (Accumulation$\setminus$Distribution) Oscillator \\
		CCI & Commodity Channel Index \\
	\end{tabular}
	\caption{Technical Indicators}
	\label{table: technical}
\end{table}

\subsection{Results}
In this section, results of five different experiments are explained. Since one of the baseline algorithms uses PCA for dimension reduction, the performance of the algorithm with different number of principal components is tested. In order to make the situation equal for the other baseline algorithms, these algorithms are tested several times with the same condition. Then, average F-measure of the algorithms are compared. More details about used notations are in Table \ref{table:algorithms}. 

\begin{table}
	\centering
	\begin{tabular}{c| c}
		
		Algorithm & Explanation \\ 
		\hline \hline
		3D-CNNpred & Our method\\
		2D-CNNpred & Our method\\
		PCA+ANN {\citep{zhong2017forecasting}} & PCA as dimension reduction and ANN as classifier  \\ 
		Technical {\citep{kara2011predicting}} & Technical indicators and ANN as classifier  \\
		CNN-cor {\citep{gunduz2017intraday}} & A CNN with mentioned structure in the paper \\
	\end{tabular}
	\caption{Description of used algorithms}
	\label{table:algorithms}
\end{table}

Tables [6-10] summarize the results for the baseline algorithms as well as our suggested models on S\&P 500 index, Dow Jones Industrial Average, NASDAQ Composite, NYSE Composite and RUSSELL 2000 historical data.  Each table consists of different statistical information about a specific market. The results include average of F-measure, as well as the best F-measure and the standard deviation of F-measures for the predictions in different runs.  Standard deviation of produced F-measures demonstrates how much generated results of a model vacillates over their mean. Models with lower standard deviation are more robust. Also P-values against 2D-CNNpred and 3D-CNNpred are also reported to show whether the differences are significant or not. 

\begin{table}
	\centering
	\begin{scriptsize}
	\begin{tabular}[c]{c| c| c | c |c |c }	
	Measure \textbackslash Model & Technical & CNN-cor & PCA+ANN & 2D-CNNpred & 3D-CNNpred  \\
	\hline \hline
	\specialcell{Mean of\\ F-measure} &0.4469 &0.3928  & 0.4237 & 0.4799 & 0.4837\\
	\specialcell{Best of \\F-measure} &0.5627 &0.5723  & 0.5165 & 0.5504 & 0.5532\\
	\specialcell{Standard deviation\\ of F-measure} &0.0658 &0.1017  & 0.0596 & 0.0273 & 0.0343\\
	\specialcell{P-value against \\2D-CNNpred} &0.0056 &less than 0.0001  & less than 0.0001 & 1  & 0.5903\\
	\specialcell{P-value against\\ 3D-CNNpred} &0.003 &less than 0.0001  & less than 0.0001 & 0.5903 & 1\\
	\end{tabular}
	\caption{Statistical information of S\&P 500 index using different algorithms}
	\label{table:S&P}
	\end{scriptsize}
\end{table}

\begin{table}
	\centering
	\begin{scriptsize}
		\begin{tabular}[c]{c| c| c | c |c |c }	
			Measure \textbackslash Model & Technical & CNN-cor & PCA+ANN & 2D-CNNpred & 3D-CNNpred  \\
			\hline \hline
			\specialcell{Mean of\\ F-measure} &0.415 &0.39  & 0.4283 & 0.4822 & 0.4925\\
			\specialcell{Best of \\F-measure} &0.5518 &0.5253  & 0.5392 & 0.5678 & 0.5778\\
			\specialcell{Standard deviation\\ of F-measure} &0.0625 &0.0939  & 0.064 & 0.0321 & 0.0347\\
			\specialcell{P-value against \\2D-CNNpred} &less than 0.0001 &less than 0.0001  & less than 0.0001 & 1  & 0.1794\\
			\specialcell{P-value against\\ 3D-CNNpred} &less than 0.0001 &less than 0.0001  & less than 0.0001 & 0.1794 & 1\\
		\end{tabular}
		\caption{Statistical information of Dow Jones Industrial Average index using different algorithms}
		\label{table:DJI}
	\end{scriptsize}
\end{table}

\begin{table}
	\centering
	\begin{scriptsize}
		\begin{tabular}[c]{c| c| c | c |c |c }	
			Measure \textbackslash Model & Technical & CNN-cor & PCA+ANN & 2D-CNNpred & 3D-CNNpred  \\
			\hline \hline
			\specialcell{Mean of\\ F-measure} &0.4199 &0.3796  & 0.4136 & 0.4779 & 0.4931\\
			\specialcell{Best of \\F-measure} &0.5487 &0.5498  & 0.5312 & 0.5219 & 0.5576\\
			\specialcell{Standard deviation\\ of F-measure} &0.0719 &0.1114  & 0.0553 & 0.0255 & 0.0405\\
			\specialcell{P-value against \\2D-CNNpred} &less than 0.0001 &less than 0.0001  & less than 0.0001 & 1  & 0.0509\\
			\specialcell{P-value against\\ 3D-CNNpred} &less than 0.0001 &less than 0.0001  & less than 0.0001 & 0.0509 & 1\\
		\end{tabular}
		\caption{Statistical information of NASDAQ Composite index using different algorithms}
		\label{table:NASDAQ}
	\end{scriptsize}
\end{table}

\begin{table}
	\centering
	\begin{scriptsize}
		\begin{tabular}[c]{c| c| c | c |c |c }	
			Measure \textbackslash Model & Technical & CNN-cor & PCA+ANN & 2D-CNNpred & 3D-CNNpred  \\
			\hline \hline
			\specialcell{Mean of\\ F-measure} &0.4071 &0.3906  & 0.426 & 0.4757 & 0.4751\\
			\specialcell{Best of \\F-measure} &0.5251 &0.5376  & 0.5306 & 0.5316 & 0.5592\\
			\specialcell{Standard deviation\\ of F-measure} &0.0556 &0.0926  & 0.059 & 0.0314 & 0.0384\\
			\specialcell{P-value against \\2D-CNNpred} &less than 0.0001 &less than 0.0001  & less than 0.0001 & 1  & 0.9366\\
			\specialcell{P-value against\\ 3D-CNNpred} &less than 0.0001 &less than 0.0001  & less than 0.0001 & 0.9366 & 1\\
		\end{tabular}
		\caption{Statistical information of NYSE Composite using different algorithms}
		\label{table:NYA}
	\end{scriptsize}
\end{table}

\begin{table}
	\centering
	\begin{scriptsize}
		\begin{tabular}[c]{c| c| c | c |c |c }	
			Measure \textbackslash Model & Technical & CNN-cor & PCA+ANN & 2D-CNNpred & 3D-CNNpred  \\
			\hline \hline
			\specialcell{Mean of\\ F-measure} &0.4525 &0.3924  & 0.4279 & 0.4775 & 0.4846\\
			\specialcell{Best of \\F-measure} &0.5665 &0.5602  & 0.5438 & 0.532 & 0.5787\\
			\specialcell{Standard deviation\\ of F-measure} &0.0655 &0.0977  & 0.066 & 0.0271 & 0.0371\\
			\specialcell{P-value against \\2D-CNNpred} &0.0327 &less than 0.0001  & 0.0001 & 1  & 0.3364\\
			\specialcell{P-value against\\ 3D-CNNpred} &0.01 &less than 0.0001  & less than 0.0001 & 0.3364 & 1\\
		\end{tabular}
		\caption{Statistical information of RUSSELL 2000 using different algorithms}
		\label{table:RUSSELL}
	\end{scriptsize}
\end{table}

To summarize and compare the performance of different algorithms, average results of them in 5 market indices are shown in Fig \ref{fig: final-results}.

\begin{figure}[!h]
	\makebox[\textwidth]{\includegraphics[width=0.7\paperwidth]{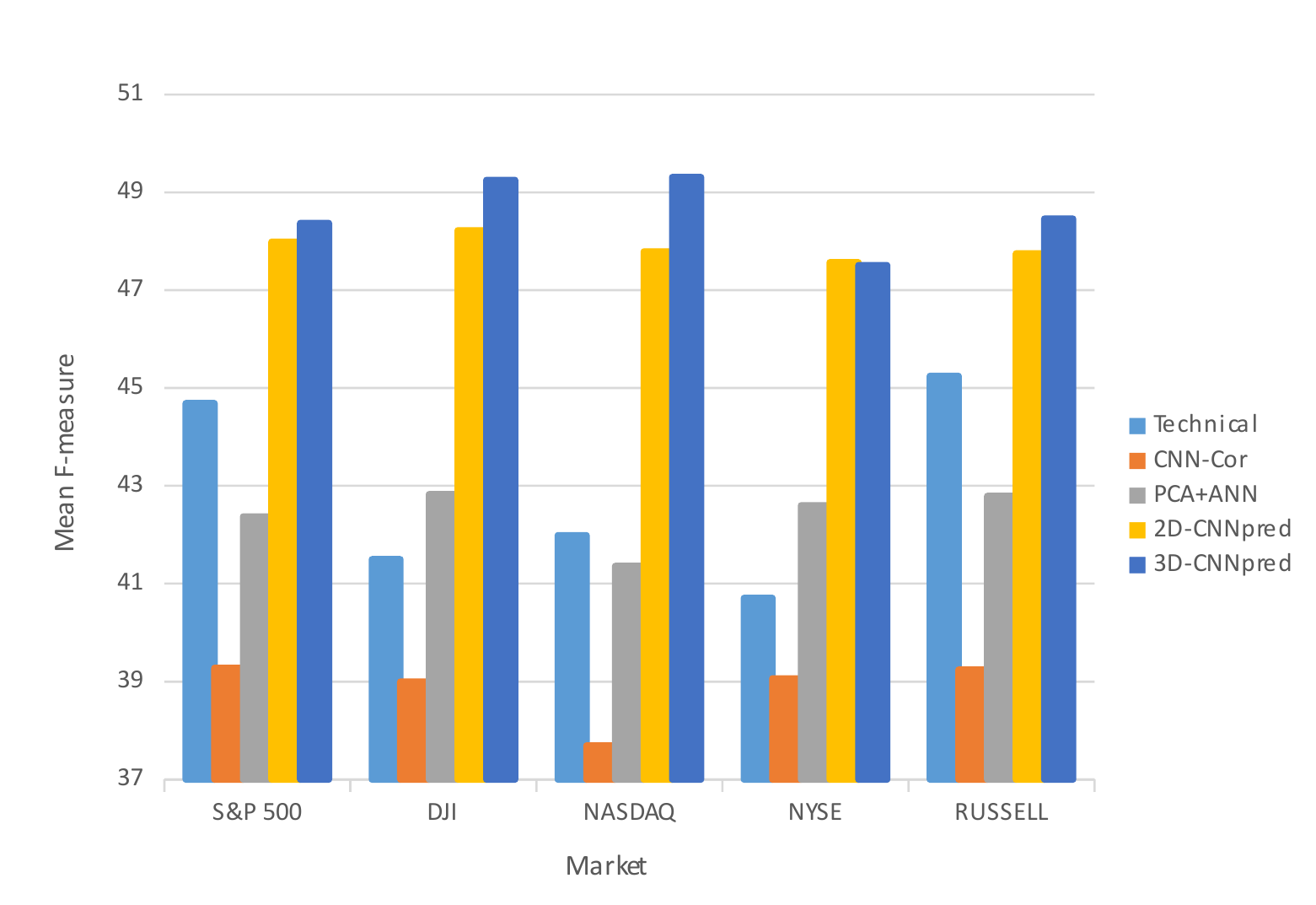}}
	\caption{Mean F-measure of different algorithms in different markets}
	\label{fig: final-results}
\end{figure}

\section{Discussion}
It is obvious from the results that both 2D-CNNpred and 3D-CNNpred statistically outperformed the other baseline algorithms. The difference between F-measure of our model and baseline algorithm which uses only ten technical indicators is obvious. A plausible reason for that could be related to the insufficiency of technical indicators for prediction as well as using a shallow ANN instead of our deep prediction model. Even, more initial features and incorporation of PCA which is a famous feature extraction algorithm did not improve the results as it was expected. A drawback of these two approaches may be the fact that they use shallow ANN that has only one hidden layer and its ability in feature extraction is limited. It demonstrates that adding more basic features is not enough by itself without improving the feature extraction algorithm. Our framework has two advantages over these two baseline algorithm that have led to its superiority in performance: First, it uses a rich set of feature containing useful information for stock prediction. Second, it uses a deep learning algorithm that extracts sophisticated features out of primary ones. 

The next baseline algorithm was CNN-Cor which had the worst results among all the tested algorithms. CNN’s ability in feature extraction is highly dependent on wisely selection of its parameters in a way that fits the problem for which it is supposed to be applied. With regards to the fact that both 2D-CNNpred and CNN-Cor used the same feature set and they were trained almost in the same way, poor results of CNN-Cor compared to 2D-CNNpred is possibly the result of the design of the 2D-CNN. Generally, the idea of using $3\times 3$ and $5\times 5$ filters seems skeptical. The fact that these kinds of filters are popular in computer vision does not guarantee that they would work well in stock market prediction as well. In fact, prediction with about 9\% lower F-measure on average in comparison to the 2D-CNNpred showed that designing the structure of CNN is the core challenge in applying CNNs for stock market prediction. A poorly designed CNN can adversely influence the results and make CNN’s performance even worse than a shallow ANN. 

Finally, an advantage of 3D-CNNpred over 2D-CNNpred which could be the reason for slightly better performance of 3D-CNN is that the latter one can combine information from different markets into a high-level feature while 2D-CNNpred has access to one market’s initial features in its feature extraction phase.

\section{Conclusion}
The noisy and nonlinear behavior of prices in financial markets makes prediction in those markets a difficult task. A better prediction can be gained by having better features. In this paper, we tried to use a wide collection of information, including historical data from the target market, general economic data and information from other possibly correlated stock markets. Also, two variations of a deep CNN-based framework were introduced and applied to extract higher-level features from that rich set of initial features. 

The suggested framework, CNNpred, was tested to make predictions in S\&P 500, NASDAQ, DJI, NYSE, and RUSSELL. Final results showed the significant superiority of two versions of CNNPred over the state of the art baseline algorithms. CNNpred was able to improve the performance of prediction in all the five indices over the baseline algorithms by about 3\% to 11\%, in terms of F-measure. In addition to confirming the usefulness of the suggested approach, these observations also suggest that designing the structures of CNNs for the stock prediction problems is possibly a core challenge that deserves to be further studied. 

\clearpage

\section*{Appendix I. Description of features}
\small
The list of features from different categories used as initial feature set representing each sample:
\begin{tiny}
\begin{longtable}[c]{c| c | c |c |c }
		
	\# & Feature & Description & Type & Source / Calculation  \\
	\hline \hline
	\endhead
	1 & Day & which day of week & Primitive & Pandas \\
	2 & Close & Close price & Primitive & Yahoo Finance \\
	3 & Vol & Relative change of volume & Technical Indicator & TA-Lib \\
	%4 & MOM & Return of 1 day before & Technical Indicator & TA-Lib \\
	4 & MOM-1 & Return of 2 days before & Technical Indicator & TA-Lib \\
	5 & MOM-2 & Return of 3 days before & Technical Indicator & TA-Lib \\
	6 & MOM-3 & Return of 4 days before & Technical Indicator & TA-Lib \\
	7 & ROC-5 & 5 days Rate of Change & Technical Indicator & TA-Lib \\
	8 & ROC-10 & 10 days Rate of Change & Technical Indicator & TA-Lib \\
	9 & ROC-15 & 15 days Rate of Change & Technical Indicator & TA-Lib \\
	10 & ROC-20 & 20 days Rate of Change & Technical Indicator & TA-Lib \\
	11 & EMA-10 & 10 days Exponential Moving Average & Technical Indicator & TA-Lib \\
	12 & EMA-20 & 20 days Exponential Moving Average & Technical Indicator & TA-Lib \\
	13 & EMA-50 & 50 days Exponential Moving Average & Technical Indicator & TA-Lib \\
	14 & EMA-200 & 200 days Exponential Moving Average & Technical Indicator & TA-Lib \\
	15 & DTB4WK & 4-Week Treasury Bill: Secondary Market Rate & Economic & FRBSL \\
	16 & DTB3 & 3-Month Treasury Bill: Secondary Market Rate & Economic & FRBSL \\
	17 & DTB6 & 6-Month Treasury Bill: Secondary Market Rate & Economic & FRBSL \\
	18 & DGS5 & 5-Year Treasury Constant Maturity Rate & Economic & FRBSL \\
	19 & DGS10 & 10-Year Treasury Constant Maturity Rate & Economic & FRBSL \\
	20 & DAAA & Moody's Seasoned Aaa Corporate Bond Yield & Economic & FRBSL \\
	21 & DBAA & Moody's Seasoned Baa Corporate Bond Yield & Economic & FRBSL \\
	22 & TE1 & DGS10-DTB4WK & Economic & FRBSL \\
	23 & TE2 & DGS10-DTB3 & Economic & FRBSL \\
	24 & TE3 & DGS10-DTB6 & Economic & FRBSL \\
	25 & TE5 & DTB3-DTB4WK & Economic & FRBSL \\
	26 & TE6 & DTB6-DTB4WK & Economic & FRBSL \\
	27 & DE1 & DBAA-BAAA & Economic & FRBSL \\
	28 & DE2 & DBAA-DGS10 & Economic & FRBSL \\
	29 & DE4 & DBAA-DTB6 & Economic & FRBSL \\
	30 & DE5 & DBAA-DTB3 & Economic & FRBSL \\
	31 & DE6 & DBAA-DTB4WK & Economic & FRBSL \\
	32 & CTB3M & \specialcell{Change in the market yield on U.S. Treasury securities
	at \\ 3-month constant maturity, quoted on investment
	basis} & Economic & FRBSL \\
	33 & CTB6M & \specialcell{Change in the market yield on U.S. Treasury securities
	at \\ 6-month constant maturity, quoted on investment
	basis} & Economic & FRBSL \\
	34 & CTB1Y & \specialcell{Change in the market yield on U.S. Treasury securities
	at \\ 1-year constant maturity, quoted on investment
	basis} & Economic & FRBSL \\
	35 & Oil & Relative change of oil price(WTI), Oklahoma & Commodity & FRBSL \\
	36 & Oil & Relative change of oil price(Brent) & Commodity & Investing.com \\
	37 & Oil & Relative change of oil price(WTI) & Commodity & Investing.com \\
	38 & Gold & Relative change of gold price (London market) & Commodity & FRBSL \\
	39 & Gold-F & Relative change of gold price futures & Commodity & Investing.com \\
	40 & XAU-USD & Relative change of gold spot U.S. dollar & Commodity & Investing.com \\
	41 & XAG-USD & Relative change of silver spot U.S. dollar & Commodity & Investing.com \\
	42 & Gas & Relative change of gas price & Commodity & Investing.com \\
	43 & Silver & Relative change of silver price & Commodity & Investing.com \\
	44 & Copper & Relative change of copper future & Commodity & Investing.com \\
	45 & IXIC & Return of NASDAQ Composite index & World Indices & Yahoo Finance \\
	46 & GSPC & Return of S\&P 500 index & World Indices & Yahoo Finance \\
	47 & DJI & Return of Dow Jones Industrial Average & World Indices & Yahoo Finance \\
	48 & NYSE & Return of NY stock exchange index & World Indices & Yahoo Finance \\
	49 & RUSSELL & Return of RUSSELL 2000 index & World Indices & Yahoo Finance \\
	50 & HSI & Return of Hang Seng index & World Indices & Yahoo Finance \\
	51 & SSE & Return of Shang Hai Stock Exchange Composite index & World Indices & Yahoo Finance \\
	52 & FCHI & Return of CAC 40 & World Indices & Yahoo Finance \\
	53 & FTSE & Return of FTSE 100 & World Indices & Yahoo Finance \\
	54 & GDAXI & Return of DAX & World Indices & Yahoo Finance \\
	55 & USD-Y & Relative change in US dollar to Japanese yen exchange rate  & Exchange Rate & Yahoo Finance \\
	56 & USD-GBP & Relative change in US dollar to British pound exchange rate & Exchange Rate & Yahoo Finance \\
	57 & USD-CAD & Relative change in US dollar to Canadian dollar exchange rate& Exchange Rate & Yahoo Finance \\
	58 & USD-CNY & Relative change in US dollar to Chinese yuan exchange rate& Exchange Rate & Yahoo Finance \\
	59 & USD-AUD & Relative change in US dollar to Australian dollar exchange rate& Exchange Rate & Investing.com \\
	60 & USD-NZD & Relative change in US dollar to New Zealand dollar exchange rate& Exchange Rate & Investing.com \\
	61 & USD-CHF & Relative change in US dollar to Swiss franc exchange rate& Exchange Rate & Investing.com \\
	62 & USD-EUR & Relative change in US dollar to Euro exchange rate& Exchange Rate & Investing.com \\
	
	63 & USDX & Relative change in US dollar index & Exchange Rate & Investing.com \\
	
	64 & XOM & Return of Exon Mobil Corporation & U.S. Companies & Yahoo Finance \\ 
	65 & JPM & Return of JPMorgan Chase \& Co. & U.S. Companies & Yahoo Finance\\
	66 & AAPL & Return of Apple Inc. & U.S. Companies & Yahoo Finance\\
	67 & MSFT & Return of Microsoft Corporation & U.S. Companies & Yahoo Finance\\
	68 & GE & Return of General Electric Company & U.S. Companies & Yahoo Finance\\
	69 & JNJ & Return of Johnson \& Johnson & U.S. Companies & Yahoo Finance\\
	70 & WFC & Return of Wells Fargo \& Company & U.S. Companies & Yahoo Finance\\
	71 & AMZN & Return of Amazon.com Inc. & U.S. Companies & Yahoo Finance\\
	
	72 & FCHI-F & Return of CAC 40 Futures & Futures & Investing.com \\
	73 & FTSE-F & Return of FTSE 100 Futures & Futures & Investing.com \\
	74 & GDAXI-F & Return of DAX Futures & Futures& Investing.com \\
	75 & HSI-F & Return of Hang Seng index Futures & Futures & Investing.com \\
	76 & Nikkei-F & Return of Nikkei index Futures & Futures & Investing.com \\
	77 & KOSPI-F & Return of Korean stock exchange Futures & Futures & Investing.com \\
	78 & IXIC-F & Return of NASDAQ Composite index Futures & Futures& Investing.com \\
	79 & DJI-F & Return of Dow Jones Industrial Average Futures & Futures& Investing.com \\
	80 & S\&P-F & Return of S\&P 500 index Futures & Futures& Investing.com \\
	81 & RUSSELL-F & Return of RUSSELL Futures & Futures& Investing.com \\	
	82 & USDX-F & Relative change in US dollar index futures & Exchange Rate & Investing.com \\
	
	\caption{Description of used indices}
	\label{table:features}
\end{longtable}
\end{tiny}

\section*{References}

\bibliography{mybibfile}

\end{document}